# ANALYSIS OF HEART DISEASES DATASET USING NEURAL NETWORK APPROACH

Dr. K. Usha Rani

Dept. of Computer Science
Sri Padmavathi Mahila Visvavidyalayam (Women's University)
Tirupati - 517502 , Andhra Pradesh, India
usharanikuruba@yahoo.co.in

## ABSTRACT

*One of the important techniques of Data mining is Classification. Many real world problems in various fields such as business, science, industry and medicine can be solved by using classification approach. Neural Networks have emerged as an important tool for classification. The advantages of Neural Networks helps for efficient classification of given data. In this study a Heart diseases dataset is analyzed using Neural Network approach. To increase the efficiency of the classification process parallel approach is also adopted in the training phase.*

## KEYWORDS

*Data mining, Classification, Neural Networks, Parallelism, Heart Disease*

## 1. INTRODUCTION

Data mining is the process of automating information (knowledge) discovery. Knowledge Discovery in Databases (KDD) is the process of getting high-level knowledge from low-level data. Data mining plays an important role in the KDD. Data mining is an interdisciplinary field. Its main aim is to uncover relationships in data and to predict outcomes. Researchers are trying to find satisfactory solutions in a reasonable time through search techniques as many problems are difficult to be solved in a feasible time by analytically. Hence data mining got its importance.

Data mining helps to extract patterns in the process of knowledge discovery in databases in which intelligent methods are applied. The emerging field of data mining promises to provide new techniques and intelligent tools which help the human to analyze and understand large bodies of data remains on difficult and unsolved problem. The common functions in current data mining practice include Classification, Regression, Clustering, Rule generation, Discovering association rules, summarization, dependency modeling, and sequence analysis. Classification is one of the important techniques of data mining. The input to the classification problem is a data-set called the training set having a number of attributes. The attributes are either continuous or categorical. One of the categorical attributes is class label or the classifying attribute. The objective is to use the training set to build a model of the class label based on the other attributes such that the model can be used to classify new data not from the training data-set.

Various data mining problems can be handled effectively by soft computing techniques. These techniques are fuzzy logic, neural networks, genetic algorithms and rough sets,
which will lead to an intelligent, interpretable, low cost solution than traditional techniques. Artificial Neural Network (ANN) is one of the most used data mining method to extract patterns in an intelligent and reliable way and has been greatly used to find models that describe data relationship [1-2]. In view of the above said significant characteristics of ANN, Neural Network technique is adopted in this study for data classification. Parallel processing is implemented at each node in different layers in the network.

Data mining techniques have been widely used in diagnostic and health care applications because of their predictive power. Data mining algorithms can learn from past examples in clinical data and model the oftentimes non-linear relationships between the independent and dependent variables. The resulting model represents formalized knowledge, which can often provide a good diagnostic opinion.

In this study the neural network approach to generate efficient classification rules is proposed. To perform classification task of medical data, the neural network is trained using Back propagation algorithm. As the structure of neural network is convenient for parallel processing, the output at each neuron in different layers is calculated in parallel. The performance of the network is analyzed with various types of test data.

The overall organization of the paper is as follows. After the introduction fundamental issues of neural networks for classification are presented in Section 2. Section 3 presents the applications of neural networks in the medical field. In Section 4 a neural network approach to generate efficient classification rules is proposed and the performance of the network is analyzed. Finally, Section 5 concludes the paper.

## 2. Neural Networks

A Neural Network (NN) consists of many Processing Elements (PEs), loosely called "neurons" and weighted interconnections among the PEs. Each PE performs a very simple computation, such as calculating a weighted sum of its input connections, and computes an output signal that is sent to other PEs. The training (mining) phase of a NN consists of adjusting the weights (real-valued numbers) of the interconnections, in order to produce the desired output [5].

The Artificial Neural Network (ANN) is a technique that is commonly applied to solve data mining applications. Neural Network is a set of processing units when assembled in a closely interconnected network, offers rich structure exhibiting some features of the biological neural network. The structure of neural network provides an opportunity to the user to implement parallel concept at each layer level. Another significant characteristic of ANN is fault tolerance. ANNs are well suited in situations where information is noisy and uncertain. ANN are an information processing methodology that differs drastically from conventional methodologies in that it employ training by examples to solve problem rather than a fixed algorithm [3,4]. They can be divided into two types based on the training method: Supervised training and Unsupervised training. Networks that are supervised require the actual desired output for each input where as unsupervised networks does not require the desired output for each input.

A key feature of neural networks is an iterative learning process in which data cases are presented to the network one at a time, and the weights associated with the input values are adjusted each time [5]. After all cases are presented, the process often starts over again. During this learning phase, the network learns by adjusting the weights so as to be able to predict the correct class label of input samples. Once a network has been structured for a particular application, that network is ready to be trained. To start this process, the initial weights are chosen randomly. Then the training or learning, begins.

The most popular neural network algorithm is back-propagation algorithm. Although many types of neural networks can be used for classification purposes [6], the focus is on the feedforward multilayer networks or multilayer perceptrons which are the most widely studied and used neural network classifiers. The feedforward, back-propagation architecture was developed in the early 1970's. This back-propagation architecture is the most popular, effective, and easy-to-learn model for complex, multi-layered networks. Its greatest strength is in non-linear solutions to ill-defined problems. The typical back-propagation network has an input layer, an output layer, and at least one hidden layer. There is no theoretical limit on the number of hidden layers but typically there are just one or two. Some work has been done which indicates that a maximum of five layers (one input layer, three hidden layers and an output layer) are required to solve problems of any complexity. Each layer is fully connected to the succeeding layer.

Training inputs are applied to the input layer of the network, and desired outputs are compared at the output layer. During the learning process, a forward sweep is made through the network, and the output of each element is computed layer by layer. The difference between the output of the final layer and the desired output is back-propagated to the previous layers, usually modified by the derivative of the transfer function, and the connection weights are normally adjusted. This process proceeds for the previous layers until the input layer is reached [7].

The advantages of Neural Networks for classification are:

- Neural Networks are more robust because of the weights
- The Neural Networks improves its performance by learning.  This may continue even after the training set has been applied.
- The use of Neural Networks can be parallelized as specified above for better performance.
- There is a low error rate and thus a high degree of accuracy once the appropriate training has been performed.
- Neural Networks are more robust in noisy environment

## 3.  NEURAL NETWORKS IN THE MEDICAL field

Neural networks are known to produce highly accurate results in practical applications.  Neural networks have been successfully applied to a variety of real world classification tasks in industry, business and science [8]. Also they have been  applied to various areas of medicine, such as diagnostic aides, medicine, biochemical analysis, image analysis, and drug development**.** They are used in the analysis of medical images from a variety of imaging modalities. Applications in this area include tumor detection in ultra-sonograms, detection and classification of micro calcifications in mammograms, classification of chest x-rays, and tissue and vessel classification in Magnetic Resonance Images. Artificial neural networks provide a powerful tool to help doctors analyze, model, and make sense of complex clinical data across a broad range of medical applications [9-21].

As the volume of stored data increases, data mining techniques assume an  important role in finding patterns and extracting knowledge to provide better patient care and effective diagnostic capabilities.  Neural networks can be used to extract rules from a disease classification. From the rules system so discovered, we can predict if someone will have a particular stage of a particular disease.

## 4. EXPERIMENT - CLASSIFICATION OF HEART DISEASES DATASET

In this experiment the medical data related to Heart diseases is considered.  This dataset was obtained from Cleveland database. This is publicly available dataset in the Internet. Cleveland dataset concerns classification of person into normal and abnormal person regarding heart diseases.

**Data Representation:**

Number of instances: 414.

Number of attributes : 13 and a class attribute

Class:

    Class0: Normal Person.

    Class1: first stroke

    Class2: second stroke

    Class3: end of life

**Attribute Description :**

| Attribute | Description | Range |
|---|---|---|
| Age | Age in years | Continuous |
| Sex | (1=male; 0=female) | 0,1 |
| Cp | --Value 1:typical angina | 1,2,3,4 |
|  | --Value 2: atypical anginal |  |
|  | --Value 3: non-anginal pain |  |
|  | --Value 4: asymptotic |  |
| trestbps | Resting blood pressure(in mm Hg) | Continuous |
| chol | Serum cholesterol in mg/dl | Continuous |
| fbs | (Fasting blood sugar .120mg/dl) | 0,1 |
|  | (1=true; 0=false) |  |
| restecg | electrocardiography results | 0,1,2 |
|  | --Value 0: normal |  |
|  | --Value 1:having ST-T wave abnormality |  |
|  | (T wave inversions and/or ST |  |
|  | Elevation or depression of>0.05mV) |  |
|  | --Value 2:showing probable or definite left |  |
|  | ventricular Hypertrophy by Estes' criteria |  |
| Thalach | Maximum heart rate achieved | Continuous |
| Exang | Exercise induced angina(1=yes;0=no) | 0,1 |
| Oldpeak | ST depression induced by exercise |  |
|  | relative to rest | Continuous |
| Slope | The slope of the peak exercise |  |
|  | ST segment | 1,2,3 |
|  | Value 1: up sloping |  |
|  | Value 2: flat |  |
|  | Value 3:down sloping |  |
| Ca | Number of major vessels (0-3) |  |
|  | Colored by fluoroscopy | Continuous |
| Thal | Normal, fixed defect, reversible defect | 3,6,7 |

**Linear Data Scaling**

Here each value is converted into the range between 0 to 1 using the following formulae

$$\text{Delta} = X_{max} - X_{min}$$
$$Y = \text{intercept } C = (X-X_{min})/\text{Delta}$$
$$\text{Slope} = m = 1/\text{Delta}$$

So we calculate Y for a given X as $Y = mX+C$.

## 4.1 Training the Neural Network

In this experiment the neural network is trained with Heart Diseases database by using feed forward neural network model and backpropagation learning algorithm with momentum and variable learning rate. The input layer of the network consists of 13 neurons to represent each attribute as the database consists of 13 attributes. The number of classes are four: 0 – normal person, 1- first stroke, 2- second stroke and 3- end of life. The output layer consists two neurons to represent these four classes. The description of the backpropagation algorithm is specified in the above is used to train the neural network during the training process. Several neural networks are constructed with and without hidden layers, i.e, single and multi layer networks and trained with heart disease dataset. Relationship between the number of epochs and the sum of squares of errors during training process for various networks can be observed from the Figures 1-2.

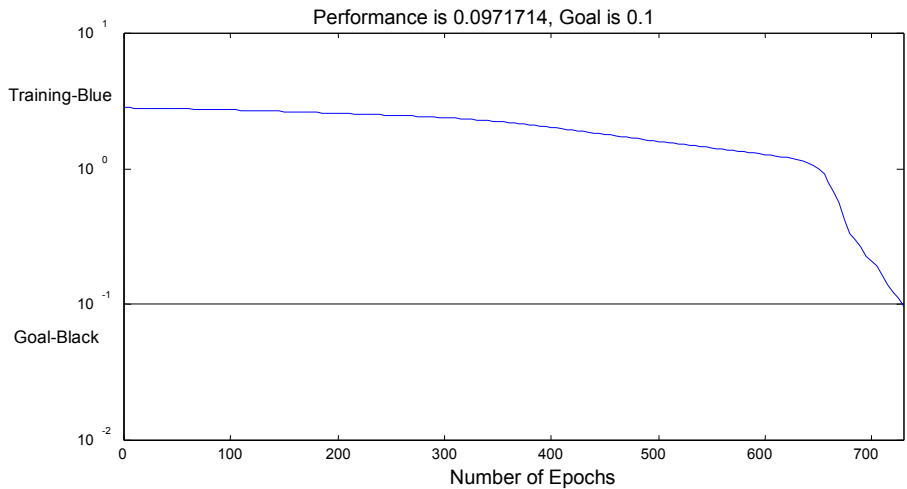

Figure 1. Training the Single Layer Network with Heart Diseases Dataset

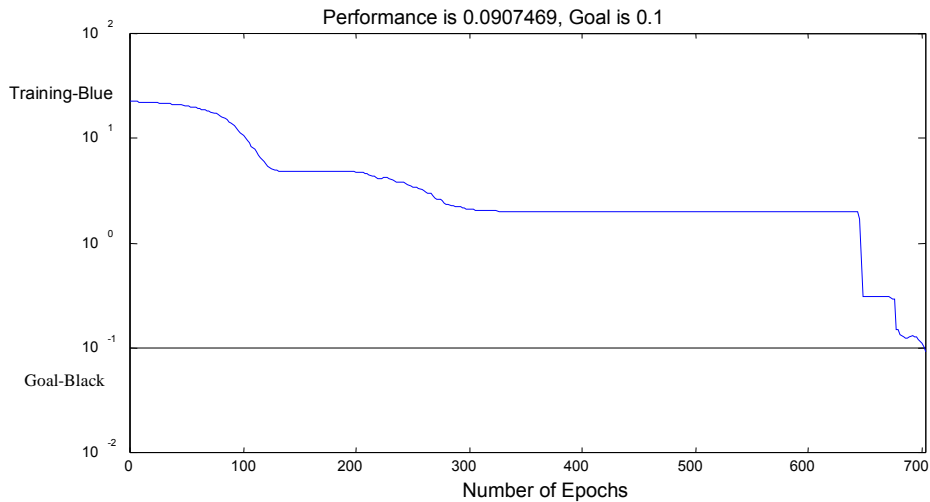

Figure 2. Training the Multi Layer Network with Heart Diseases Dataset

## 4.2 Performance of the Network

For testing the performance of the net various samples are collected as test data. The test data is given as the input to the trained network and the output of the net is calculated with the adjusted weights. The output of the net is compared with the target output to study the learning ability of the network for classifying the heart disease data. The results are tabulated in Table1.

Table 1: Experimental Results of Heart Diseases Dataset

| Training Samples | Test Samples | Classification Efficiency | |
|---|---|---|---|
| | | Single Layer | Multi Layer |
| 100 | 300 | 76% | 82% |
| 150 | 200 | 79.4% | 83% |
| 250 | 150 | 86.2% | 89.3% |
| 350 | 100 | 90.6% | 94% |

## 5 CONCLUSION

Classification is an important problem in the rapidly emerging field of data mining. Many problems in business, science, industry, and medicine can be treated as classification problems. Owing to the wide range of applicability of ANN and their ability to learn complex and non-linear relationships including noisy or less precise information, neural networks are well suited to solve problems in biomedical engineering. In this study Neural network technique is adopted for classification of medical dataset. The experiment is conducted with Heart Disease dataset by considering the single and multilayer neural network modes. Backpropogation algorithm with momentum and variable learning rate is used to train the networks. To analyze performance of the network various test data are given as input to the network. Parallelism is implemented at each neuron in all hidden and output layers to speed up the learning process. The experimental results proved that neural networks technique provides satisfactory results for the classification task.

## REFERENCES

[1] John Shafer, Rakesh Agarwal, and Manish Mehta, (1996) "SPRINT:A scalable parallel classifier for data mining", In Proc. Of the VLDB Conference, Bombay, India..

[2] Sunghwan Sohn and Cihan H. Dagli, (2004) "Ensemble of Evolving Neural Networks in classification", Neural Processing Letters 19: 191-203, Kulwer Publishers.

[3] K. Anil Jain, Jianchang Mao and K.M. Mohiuddi, (1996) "Artificial Neural Networks: A Tutorial", IEEE Computers, pp.31-44.

[4] George Cybenk,, (1996)"Neural Networks in Computational Science and Engineerin", IEEE Computational Science and Engineering, pp.36-42

[5] R. Rojas, (1996) "Neural Networks: a systematic introduction", Springer-Verlag.

[6] R.P.Lippmann,"Pattern classification using neural networks, (1989)" IEEE Commun. Mag., pp. 47–64.

[7] Simon Haykin, (2001) "Neural Networks – A Comprehensive Foundation", Pearson Education.

[8] B.Widrow, D. E. Rumelhard, and M. A. Lehr, (1994) "Neural networks: Applications in industry, business and science," *Commun. ACM*, vol. 37, pp.93–105.

[9] W. G. Baxt, (1990) "Use of an artificial neural network for data analysis in clinical decision-making: The diagnosis of acute coronary occlusion," *Neural Comput.*, vol. 2, pp. 480–489..

[10] Dr. A. Kandaswamy, (1997) "Applications of Artificial Neural Networks in Bio Medical Engineering", The Institute of Electronics and Telecommunicatio Engineers, Proceedings of the Zonal Seminar on Neural Networks, Nov 20-21.

[11] A. Kusiak, K.H. Kernstine, J.A. Kern, K A. McLaughlin and T.L. Tseng, (2000) "Data mining: Medical and Engineering Case Studies", Proceedings of the Industrial Engineering Research Conference, Cleveland, Ohio, May21-23,pp.1-7.

[12] H. B. Burke, (1994) "Artificial neural networks for cancer research: Outcome prediction," *Sem. Surg. Oncol.*, vol. 10, pp. 73–79.

[13] H. B. Burke, P. H. Goodman, D. B. Rosen, D. E. Henson, J. N. Weinstein, F. E. Harrell, J. R. Marks, D. P. Winchester, and D. G. Bostwick, (1997) "Artificial neural networks improve the accuracy of cancer survival prediction," *Cancer*, vol. 79, pp. 857–8621997.

[14] Siri Krishan Wasan1,Vasudha Bhatnagar2 and Harleen Kaur, (2006)" The impact of Data Mining Techniques on Medical Diagnostics", Data Science Journal, Volume 5, 119-126.

[15] Scales, R., & Embrechts, M., (2002) "Computational Intelligence Techniques for Medical Diagnostic", Proceedings of Walter Lincoln Hawkins, Graduate Research Conference from the World Wide Web:
http://www.cs.rpi.edu/~bivenj/MRC/proceedings/papers/researchpaper.pdf

[16] S. M. Kamruzzaman , Md. Monirul Islam, (2006)" An Algorithm to Extract Rules from Artificial Neural Networks for Medical Diagnosis Problems", International Journal of Information Technology, Vol. 12 No. 8.

[17] Hasan Temurtas, Nejat Yumusak, Feyzullah Temurtas, (2009)" A comparative study on diabetes disease diagnosis using neural networks", Expert Systems with Applications: An International Journal , Volume 36 Issue 4.

[18]     D Gil, M Johnsson, JM Garcia Chamizo, (2009) , "Application of artificial neural networks in the diagnosis of urological dysfunctions", Expert Systems with Applications Volume 36, Issue 3, Part 2, Pages 5754-5760, Elsevier

[19]     R. Dybowski and V. Gant, (2007), "Clinical Applications of Artificial Neural Networks", Cambridge University Press.

[20]     O. Er, N. Yumusak and F. Temurtas, (2010) "Chest disease diagnosis using artificial neural networks", Expert Systems with Applications, Vol.37, No.12, pp. 7648-7655.

[21]     S. Moein, S. A. Monadjemi and P. Moallem, (2009) "A Novel Fuzzy-Neural Based Medical Diagnosis System", International Journal of Biological & Medical Sciences, Vol.4, No.3, pp. 146-150.